\documentclass[Afour,sagev,times]{sagej}

\usepackage{moreverb,url}
\usepackage[T1]{fontenc}
\usepackage{times}

\usepackage{amsmath}
\usepackage{dcolumn}
\usepackage{graphics}
\usepackage{graphicx}
\usepackage{amsmath}
\usepackage{amsfonts}
\usepackage{amssymb}
\usepackage{color}

\usepackage[colorlinks,bookmarksopen,bookmarksnumbered,citecolor=red,urlcolor=red]{hyperref}

\newcommand\BibTeX{{\rmfamily B\kern-.05em \textsc{i\kern-.025em b}\kern-.08em
T\kern-.1667em\lower.7ex\hbox{E}\kern-.125emX}}

\def\volumeyear{2025}

\begin{document}

\runninghead{Cross-Domain Wi-Fi Signal Adaptation with Digital Shielding}

\title{Digital Shielding for Cross-Domain Wi-Fi Signal Adaptation using Relativistic Average Generative Adversarial Network}

\author{Danilo Avola\affilnum{1}, 
        Federica Bruni\affilnum{1}, 
        Gian Luca Foresti\affilnum{2},
        Daniele Pannone\affilnum{1},
        and Amedeo Ranaldi\affilnum{1}}

\affiliation{\affilnum{1}
Department of Computer Science, Sapienza University of Rome,\\
Via Salaria 113, 00198, Rome, Italy\\
E-mail: \{avola,bruni,pannone,ranaldi\}@di.uniroma1.it\\
\affilnum{2}~Department of Mathematics, Computer Science and Physics, University of Udine,\\
Via delle Scienze 206, 33100, Udine, Italy\\
E-mail: gianluca.foresti@uniud.it}

\corrauth{Danilo Avola\\Department of Computer Science, Sapienza University of Rome}
\email{avola@di.uniroma1.it}

\begin{abstract}
Wi-Fi sensing leverages radio-frequency signals transmitted by Wi-Fi devices to analyze environments, thus enabling tasks such as people tracking, intruder detection, gesture recognition, and many others. The recent proliferation of this technology is due to both the introduction of the IEEE 802.11bf standard that facilitates environmental monitoring activities and the need for increasingly powerful tools that can, for example, penetrate obstacles and ensure privacy. However, the effectiveness of Wi-Fi-based sensing solutions is strongly influenced by the environment in which these signals are acquired and processed. This is particularly critical when Wi-Fi signals are used to extract spatial and temporal information from the surrounding scene, as such data reflects both the structure of the environment and potential interfering sources. One of the main challenges in this field is therefore achieving robust generalization across domains, that is, ensuring consistent performance in the use of Wi-Fi signals even when sensing conditions vary significantly (e.g., from one room or building to another), without a significant drop in accuracy. In this context, this paper presents a novel deep model specifically designed to perform cross-domain adaptation of Wi-Fi signals by simulating a digital shielding mechanism. The proposed model uses a Relativistic average Generative Adversarial Network (RaGAN) to mimic the physical shielding behavior to suppress domain-specific data while preserving the integrity of the Wi-Fi signals. In the proposed RaGAN, both the generator and discriminator employ Bidirectional Long Short-Term Memory (Bi-LSTM) based architectures, thus allowing the processing of the waveform and time-dimension data of the Wi-Fi signals. An acrylic box, lined with electromagnetic field shielding fabric designed to replicate the effects of a Faraday cage, was constructed to isolate objects and allow the RaGAN to learn the effects of physical shielding. Spectra of various objects of the same size but different materials were acquired inside (i.e., domain-free) and outside (i.e., domain-dependent) the box to build an ad-hoc dataset for the training stage. Subsequently, a multi-class Support Vector Machine (SVM) was used to evaluate the overall performance of the domain adaptation model. The multi-class SVM was trained with the spectra acquired inside the box and tested with the spectra denoised by the RaGAN, achieving an accuracy of 96\%. The multi-class SVM has also provided an added value, showing high accuracy in discriminating the different materials that make up the objects in the dataset, thus offering an innovative approach for security systems aimed at determining the nature and composition of potentially dangerous objects worn by individuals.
\end{abstract}


\maketitle

\section{Introduction}
In the current literature, the term sensing refers to the process of detecting and measuring physical properties, events, and changes, as well as the behavior of people and objects, in indoor and outdoor environments using various types of sensors\cite{KERR2003299,9705087,CHEN2024114108}. These sensors convert physical stimuli, such as light, sound, temperature, motion, or visual acquisitions like depth maps or RGB video streams, into data that can be observed and analyzed to derive a wide range of meaningful information for real-time informed decision-making. In recent years, these sensing technologies have become increasingly widespread and are playing a key role in an ever-growing number of application domains. Among the various sensing technologies, one widely known for its versatile applications across different fields is visual sensing, better known in the literature as scene understanding. The systems implemented with this technology generally consist of a distributed network of RGB cameras (occasionally other types of cameras) placed in various environments; these cameras acquire video streams that feed into computer vision models to achieve a multitude of purposes, such as person re-identification\cite{6049261,10.1145/3243316,9353394,10.1145/3649900}, object tracking\cite{doi:10.1142/S0129065712500190,doi:10.1142/S0129065700000065,9736652,10.1145/3635155,10.3233/ICA-230702}, human action recognition\cite{doi:10.1142/S0129065712500281,8936339,doi:10.1142/S0129065723500028,10517892}, foreground modeling and analysis\cite{doi:10.1142/S012906571100281X,7551138,8118166,doi:10.1142/S0129065720500161,10.3233/ICA-130428}, and much more. Despite the remarkable achievements of computer vision techniques and models, many visual sensing applications still face significant challenges, such as illumination changes\cite{8970561,9761930}, background clutter\cite{9585547,9858168}, occlusions\cite{7018985,doi:10.1142/S0129065723500478}, and perspective distortions\cite{6967815,doi:10.1142/S012906572250040X}. Many of these challenges are cross-cutting across various fields. While current RGB devices remain the most suitable for spatial resolution, image quality, data richness, and noise management, they are inherently prone to the reported limitations. Consequently, recent efforts have been intensified to develop technologies capable of replacing or complementing current visual technologies, aiming to overcome, at least in part, these issues in visual tasks.

Wi-Fi sensing is an evolving technology that has already proven to be highly effective in a wide range of applications over the past two decades\cite{10.1145/3310194,8880681,10.1145/3436729,ABUHOUREYAH2024107171}, from smart home automation and security to healthcare monitoring and human-computer interaction. In recent years, Wi-Fi devices have been used not only for current monitoring activities but also as a type of ``vision'' system capable of capturing and shaping significant information to accomplish computer vision tasks through deep analysis of propagated signal spectra. For example, in Avola et al.\cite{9730862}, the authors use Wi-Fi signal information that has passed through various subjects to develop a person re-identification system capable of providing more robust and reliable biometric signatures than visual ones, which are dependent on factors like changes in lighting or clothing. Meanwhile, Wang et al.\cite{9008282} utilize Wi-Fi signals to reconstruct the skeletons of individuals, on which classical methods can then be applied to determine human body poses and movements. While there is potential to lose valuable information, such as the texture and colors that define the surface of objects, Wi-Fi sensing offers extraordinary capabilities, including analyzing the interior of solid objects, overcoming obstacles to mitigate occlusion issues, and providing stricter privacy constraints. The promising future of Wi-Fi applications in various domains is further supported by the recent establishment of the IEEE 802.11bf standard\cite{9941042}, which formalizes and standardizes Wi-Fi sensing capabilities within the existing IEEE 802.11 Wi-Fi framework. It is now evident, especially in light of the next generation of applications under development, that regardless of the specific application being addressed, signal integrity remains a critical factor for the reliability and performance of Wi-Fi sensing applications. This is especially crucial for preserving signal fidelity and extracting rich semantic information required for accurate and complex manipulations.

At the core of Wi-Fi sensing lies the analysis of Channel State Information (CSI)\cite{1412040,10.3233/ICA-2012-0400}, which provides a comprehensive representation of how a Wi-Fi signal propagates from the transmitter to the receiver through an environment. CSI captures a range of critical data, including amplitude, phase, delay spread, Doppler Frequency Shift (DFS), and multipath effects. These characteristics are essential for in-depth signal analysis and the implementation of advanced Wi-Fi sensing applications. However, environmental variations present a fundamental limitation to the robustness of Wi-Fi sensing systems, as they can induce significant fluctuations in CSI characteristics. As highlighted in the study by Chen et al.\cite{chenCrossDomainWiFiSensing2023}, CSI is inherently sensitive to surrounding environmental factors, including room layout, furniture placement, and static reflectors. Even minor modifications, such as repositioning objects or the presence of transient obstacles, can alter the multipath propagation of Wi-Fi signals, causing domain shifts that negatively impact sensing accuracy. These variations impact key CSI metrics, such as amplitude, phase, and DFS, leading to inconsistencies in signal representation. In such conditions, models designed to perform specific sensing tasks often fail to generalize effectively when deployed elsewhere. This is particularly problematic for pre-trained models, which tend to experience a noticeable drop in performance when applied across domains characterized by different signal propagation conditions. As a result, Wi-Fi sensing systems frequently require continuous recalibration, retraining, or robust domain adaptation strategies to remain reliable in dynamic or previously unseen environments.

Considering the points discussed so far, this paper introduces an innovative deep model specifically designed to emulate the physical effects of a Faraday cage with the aim of filtering signals and removing any domain-dependent components together with possible noises and interferences that could be present in the environment. In other words, the proposed architecture purifies signals as if they had been acquired in a completely neutral and interference-free environment. To achieve this, a model based on the Relativistic average Generative Adversarial Network (RaGAN) was designed, which digitally mimics the protective characteristics of a Faraday cage, thus preserving the integrity and informative content of the Wi-Fi signal. In this proposed version of RaGAN, the generator and discriminator are implemented using Bidirectional Long Short-Term Memory (Bi-LSTM) networks. This configuration allows the model to effectively manage and interpret the sequential data present in Wi-Fi signals. The Bi-LSTM networks process the signal in two stages: first, by analyzing the sequence in the forward direction, thus capturing the dependencies as the signal evolves over time; then, by processing the sequence in the backward direction, thus ensuring that the model also considers future dependencies and provides a more comprehensive understanding of the temporal relationships within the waveform and time-series data of the Wi-Fi signals. To generate the data needed to train the RaGAN model, an acrylic box was constructed and lined with electromagnetic field shielding fabric designed to replicate the effects of a Faraday cage. Using this box, we collected an ad-hoc dataset in which various objects of the same shape and size but composed of different materials were acquired both inside the box, i.e., neutral environment, and outside the box, i.e., with real-world interference and domain-dependent. These acquisitions were then used to teach the RaGAN model to recognize the spectral representation with and without environment-dependent data, thus enabling the network to learn how to transform the spectrum of an object affected by the environment and possible interferences back to that of the same environment-free object. To evaluate the effectiveness of the obtained results, a multi-class Support Vector Machine (SVM) was subsequently employed. The multi-class SVM was trained using the spectra acquired inside the shielded box and tested with the spectra denoised by the RaGAN, achieving an accuracy of 96\%. As an additional outcome of the research presented in this paper, which focuses on Wi-Fi signal cross-domain adaptation, the application of a multi-class classifier to verify the performance of the RaGAN also revealed the ability to classify the specific nature of the signal, successfully distinguishing whether a spectrum originated from an object composed of one material (e.g., aluminum) or another material (e.g., copper). In conclusion, the main contributions of the proposed work can be summarized as follows:
\begin{itemize}
\item To the best of our knowledge, this is the first work in the literature to propose a RaGAN-based architecture specifically designed to emulate the physical shielding effect of a Faraday cage, with the goal of effectively suppressing environment-specific information from Wi-Fi signals. The experimental results confirm the effectiveness of the proposed approach.
\item Given that the proposed method is designed to transform a signal affected by any type of interference and domain-specific data into its ideal form, as if it were acquired in an interference-free environment, this model does not rely on knowing or adapting to the specific nature of the interference or environment. It thus represents an initial step toward effective cross-domain adaptation of Wi-Fi sensing systems.
\item A one-of-a-kind dataset has been created, consisting of objects made from different materials but with identical dimensions and shapes, acquired both in the presence and absence of environment data. This dataset can serve as a valuable starting point for various studies in the application fields of cross-domain adaptation and material analysis.
\item The research proposed in this paper has shown that it is possible to distinguish between different materials based on their Wi-Fi signal features. This discovery opens up new possibilities in the emerging field of Wi-Fi-based material sensing, with significant implications, for example, in security-related studies such as detecting concealed objects or identifying hazardous materials.
\end{itemize}

The remainder of this paper is organized as follows. Related Work provides a comprehensive overview of prior studies addressing signal denoising across different areas. Proposed Method describes the architecture, detailing the Bi-LSTM-based generator and discriminator. Experimental Setup and Results introduces the collected dataset and discusses the results obtained, both in terms of the cross-domain generalization capability of the proposed method and the ability of the classifier to distinguish between different materials. Finally, Conclusion concludes the paper and outlines our ongoing research directions in Wi-Fi sensing.

\section{Related Work}
\label{related_work}
Signal noise has consistently been a significant challenge in telecommunication systems due to the unpredictable nature of various noise types and their sources. Different signals and communication channels encounter distinct forms of noise, each requiring specific techniques to mitigate these disturbances. Furthermore, the term noise can refer to very different types of signal interference depending on the specific task the system is designed to perform, ranging from signal and hardware-related disturbances to environment-specific factors. In the context of Wi-Fi sensing, noise generally refers to all kinds of information gathered by the CSI that is not relevant to the specific task. Denoising techniques are typically divided into time-domain and frequency-domain approaches. The time-domain methods focus on mitigating signal noise by analyzing temporal patterns. A representative approach is median filtering, as reported by Wang et al.\cite{7458186} on fall detection, where a weighted moving average filter was employed to reduce noise caused by environmental factors such as temperature and room conditions. Another commonly used method is Principal Component Analysis (PCA), as described by Wang et al.\cite{7875148} on human activity recognition, where the first principal component, representing the highest noise variance, is discarded, while the subsequent five components are retained. The frequency-domain methods, instead, focus on reducing noise by applying filters to isolate useful signal components while attenuating unwanted noise. Common examples of these methods are represented by the work of Ali et al.\cite{10.1145/2789168.2790109} on keystroke recognition, where a low-pass filter is used, and the work of Liu et al.\cite{10.1145/2746285.2746303} on tracking vital signs during sleep, where a band-pass filter is employed. However, time-domain denoising approaches can distort the signal, thus potentially leading to the loss of high-frequency components or the removal of significant useful information. In contrast, frequency-domain denoising techniques struggle to effectively eliminate noise within the passband.

Moving into the time-frequency domain to leverage the benefits discussed above, the Discrete Wavelet Transform (DWT) has proven effective in denoising transient waveforms, such as ElectroCardioGrams (ECG) and Transient ElectroMagnetic (TEM) signals, as described in Guo et al.\cite{9785993} and Wei et al.\cite{10.1029/2020RS007135}, respectively. Although CSI signals differ in nature, they share key characteristics like non-stationarity, which DWT effectively addresses. Its noise reduction capabilities are demonstrated in the work of Fang et al.\cite{7552529}, where the authors apply the Multilevel Discrete Wavelet Transform (MDWT) to denoise and extract signal features. Recent studies have also introduced Synchrosqueezed Wavelet Transform (SWT)\cite{Amezquita-Sanchez_2015,LiZNewMethod,PEREZRAMIREZ20161} as a powerful time-frequency analysis and denoising tool. The time-frequency approach enables DWT and SWT to capture transient signal details that single-domain methods might miss, thus making it especially suited for analyzing dynamic and non-stationary waveforms.
\begin{figure*}[ht]
\centering
\includegraphics[width=0.85\textwidth]{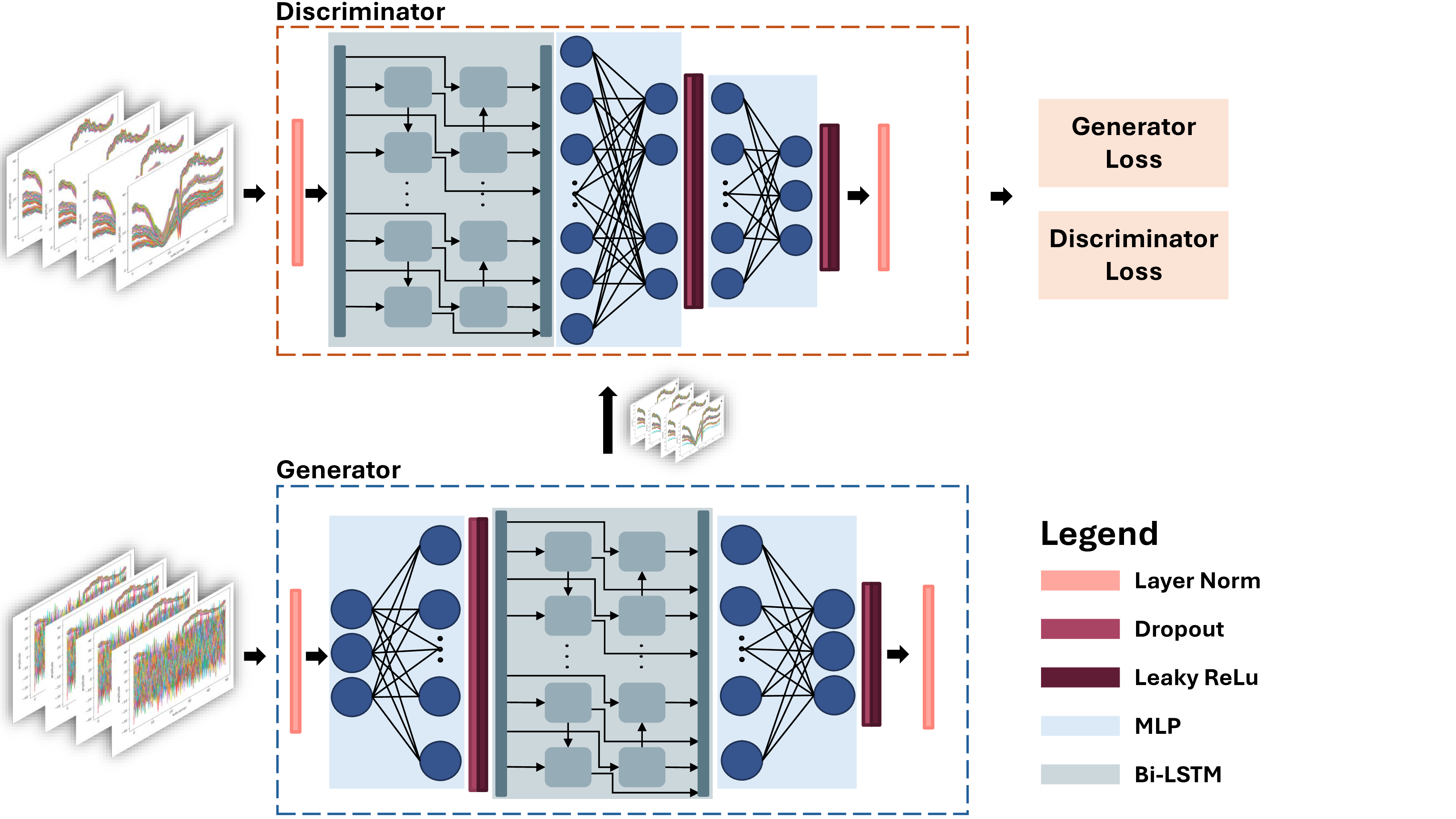}
\caption{Proposed model architecture for digital shielding. The generator and discriminator are implemented using Bi-LSTM networks to process the sequential data in Wi-Fi signals. In the first stage, the model analyzes the sequence in the forward direction to capture dependencies as the signal progresses over time; in the second, it analyzes the sequence in the backward direction to account for future dependencies. This dual approach enhances comprehension of the temporal relationships within the waveform and time-series data of Wi-Fi signals.}
\label{Figure_01}
\end{figure*}

While the techniques discussed above help reduce fluctuations and interferences caused by the environment, they still fail to eliminate domain-dependent information embedded in the CSI. To address this limitation, several cross-domain adaptation strategies have been proposed to improve the generalization capabilities of Wi-Fi sensing models. Much of the existing work in Wi-Fi sensing focuses on tasks involving moving subjects, such as activity recognition, user identification, motion detection, human tracking, and fall detection. Consequently, many studies aim to remove the static component of the CSI and retain only the dynamic part, which reflects changes due to motion. Several methods have been developed to achieve this, including the Local Extreme Value Detection (LEVD) algorithm\cite{QGesture,10.1145/3131672.3131695,10.1145/3377553}, conjugate multiplication of CSI measurements between antennas\cite{10.1145/3210240.3210314}, recursive filtering\cite{9322627,9076681}, and, to some extent, frequency-based filters, since motion-induced variations typically reside in the low-frequency range\cite{9834923}. However, while these techniques are effective in motion-related scenarios, they are not suitable for applications where the dynamic component is irrelevant, such as material identification. In such cases, a different approach is required to preserve static, domain-relevant information and suppress domain-specific noise.

In the past decade, Deep Learning (DL) methods have been applied with increasing frequency across various fields, including the area addressed in this paper\cite{10.3233/ICA-170551,10.3233/ICA-230728}. Convolutional Neural Networks (CNNs) were the first DL approaches widely adopted for denoising, particularly in the image domain. Actually, many works have approached signal denoising as an image-denoising problem. For instance, Chen et al.\cite{9258400} transformed a 1-D TEM signal into a 2D image, which was then processed by CNNs for noise reduction. Similarly, Zhu et al.\cite{8802278} applied the Short-Time Fourier Transform (STFT) to convert seismic signals from the time domain to the time-frequency domain, thus creating an image input for a denoising U-Net. Likewise, Almazrouei et al.\cite{8902756} computed the STFT of MATLAB-simulated IEEE 802.11a/n signals and fed these into a Convolutional Denoising AutoEncoder (CDAE) for effective noise reduction. Note that precise separation of noise from clear signal is not always possible and obtaining such a dataset is part of the challenge in denoising frameworks. A solution to address this issue was proposed by Wang et al.\cite{9698089}, which utilized a GAN with Wasserstein distance and Gradient Penalty (WGAN-GP) to learn the noise characteristics present in noisy TEM signals, generating noise signals to build an effective dataset for denoising models. In contrast, Yang et al.\cite{10043377} introduced an alternative approach using a Conditional GAN (CGAN) to separate crucial signal information from noise. While these works have demonstrated promising results, several limitations remain to be explored and addressed. A key challenge lies in the extended training time and slow convergence speed of current noise reduction methods based on GANs. Another limitation is ensuring the denoising method is sufficiently generalized to handle various noise types. In addition, generative model cannot remove completely the domain-specific information contained in CSI if not supported with an ad-hoc and clean dataset to learn with. Finally, maintaining the informative content within the signal after noise removal remains an essential aspect to address. Inspired by works such as that of Jolicoeur-Martineau\cite{jolicoeur-martineau2018}, which proposed the relativistic discriminator to enhance limitations in standard GANs and that of Peng et al.\cite{s23010475}, which examined the performance of a RaGAN for noise reduction in communication signals; the work proposed in this paper introduces a new paradigm for cross-domain adaptation and noise reduction based on the concept of digital shielding. This approach leverages the fast convergence of RaGAN while enabling an abstraction level in signal denoising that remains independent of the specific environment. The proposed method achieves a high degree of accuracy in cleaning, preserving carrier information and ensuring the informative content even in fine-grained classification case studies, such as distinguishing materials of different objects.

\section{Proposed Method}
\label{proposed_method}
This section presents the three key aspects of the proposed method. In the first, an explanation of the simple yet effective concept behind the proposed approach, i.e., digital shielding, is given. In the second, the implementation of the RaGAN-based architecture designed to emulate the physical effect of a Faraday cage is presented. In the last, details of the simple multi-class SVM classifier, used both to measure the performance of the RaGAN-based architecture and to validate material differentiation as further evidence of the implemented domain-independent mechanism, are shown.

\subsection{Digital Shielding in Wi-Fi Signals}
\label{digital_shielding_in_Wi-Fi_signals}
As is well known, Wi-Fi sensing covers various tasks beyond traditional communication, such as fall detection or gesture recognition. These tasks, conducted in real-world environments, are inherently susceptible to diverse  settings and contexts in which they are carried out. Indeed, Wi-Fi signals are subject to influence from numerous factors, such as electromagnetic interference, environmental morphology, and other contextual variables. The concept behind digital shielding is as follows: if a deep model can be trained to understand the behavior of a Wi-Fi signal during the execution of a specific task in an ideal, noise-free environment, then when that same signal is used in a real-world, noisy environment-dependent setting, the model should be able to clean the signal and restore it to a state similar to its ideal condition, allowing cross-domain adaptation and better generalization. A significant advantage of this approach is that the cleaning process becomes largely independent of the specific environment, as the model consistently aims to extract the fingerprint of the target signal regardless of the environment through which the signal propagates.

\subsection{Bi-LSTM-based RaGAN}
\label{bi-LSTM-based_RaGAN}
\begin{figure}
\centering
\includegraphics[width=3.2in]{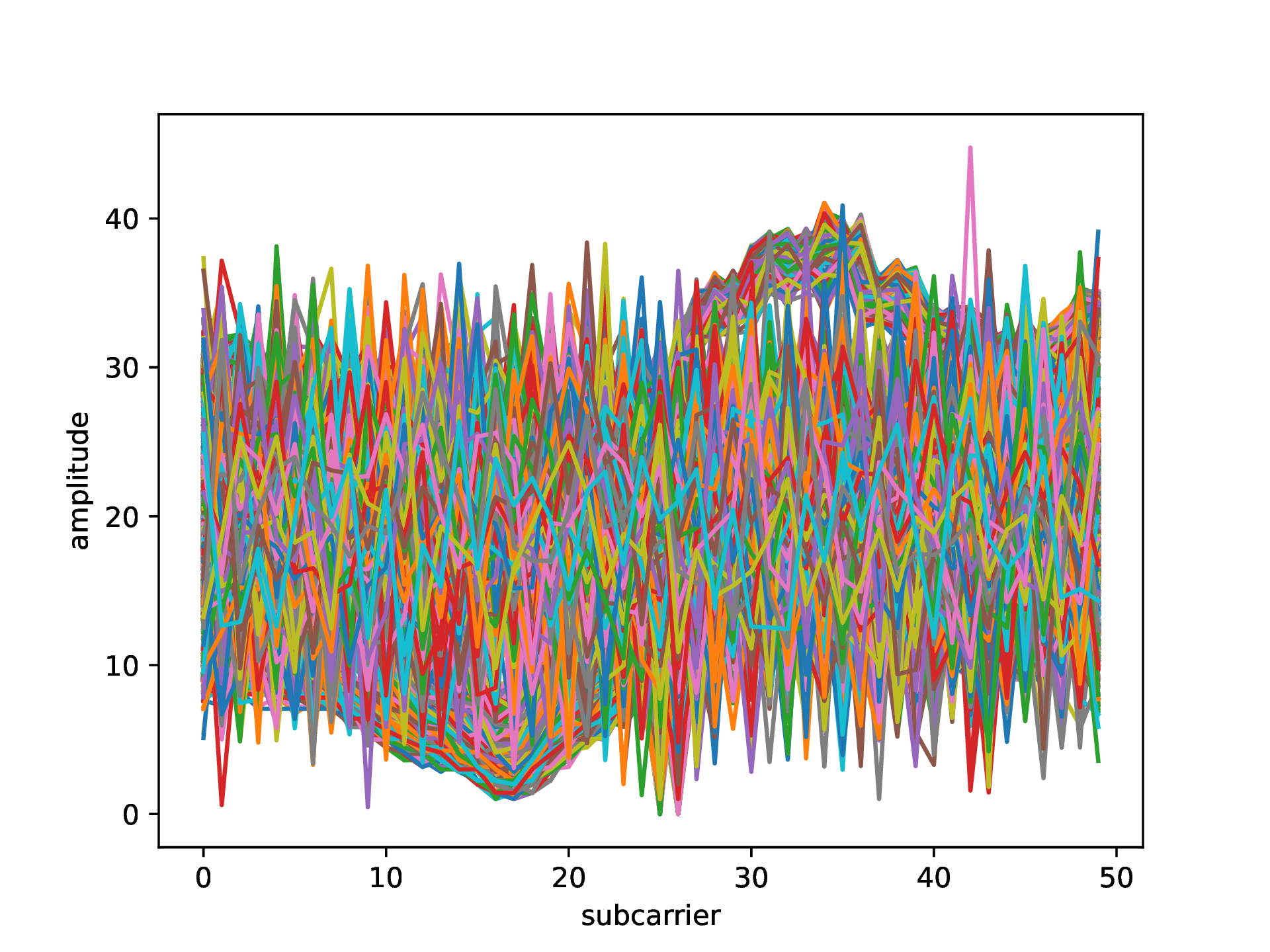}
\includegraphics[width=3.2in]{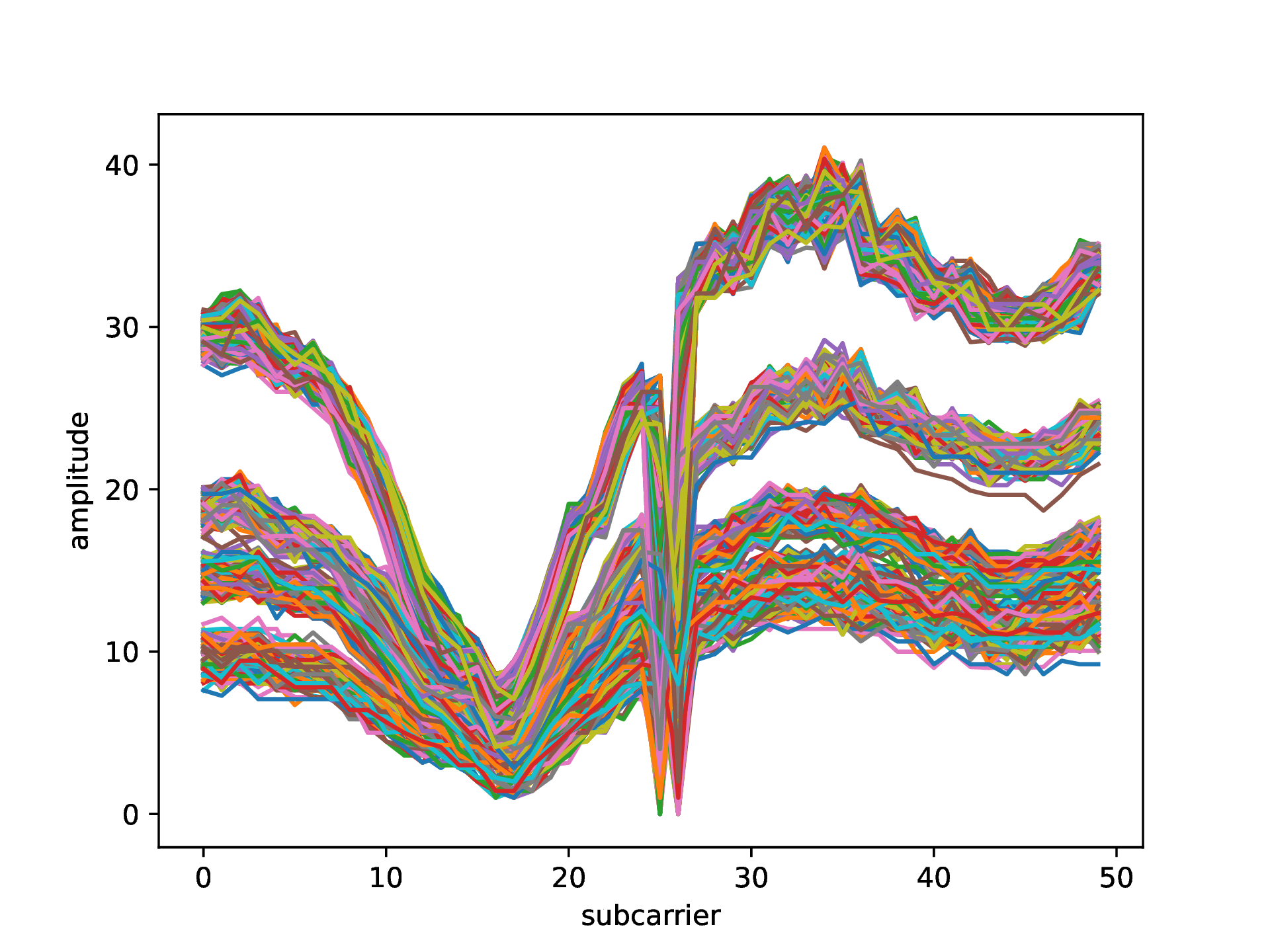}
\caption{Amplitudes extracted from the CSI: the top plot shows the acquisition of the copper cube in a real-world environment, while the bottom plot represents the acquisition inside the acrylic box lined with electromagnetic shielding fabric (i.e., Faraday cage).}
\label{Figure_02}
\end{figure}

In Fig.~\ref{Figure_01}, the proposed model architecture for digital shielding is shown. The model consists of a RaGAN, where both the generator and the discriminator are implemented using Bi-LSTM networks. This design enables the model to interpret the sequential patterns present in Wi-Fi signals. By utilizing Bi-LSTM networks, the architecture processes the signal in both the forward direction, capturing dependencies as the signal evolves over time, and the backward direction, incorporating future relationships. This bidirectional processing allows for a more comprehensive understanding of the temporal dynamics embedded within the waveform and time-series data. In order to train the network to act as a digital Faraday cage, capable of restoring real-world signals as closely as possible to their ideal representations, a specific training strategy is employed. In particular, the discriminator is trained with shielded samples, i.e., signals acquired inside an acrylic box lined with electromagnetic shielding fabric, while the generator is trained with unshielded samples, i.e., signals acquired in real-world conditions. Regarding the features extracted from the CSI, only the amplitude is used, as it encapsulates both the waveform structure and temporal characteristics of the Wi-Fi signals. As an example, Fig.~\ref{Figure_02} illustrates the amplitude computation for two acquisitions of a copper cube, performed in an unshielded (top) and shielded (bottom) environment, respectively.

\subsubsection{CSI Estimation and Amplitude Feature:}
\label{CSI_and_amplitude_feature}
details regarding the construction of the shielded box for noiseless acquisitions, the transmission and reception devices employed, and the dataset created will be provided in the experimental section. In this subsection, however, the primary focus will be on the estimation of the CSI and the subsequent extraction of the amplitude, which will serve as a fundamental feature in the proposed method to be presented in this paper.

Modern wireless communication systems rely on advanced techniques to characterize and adapt to the complex nature of propagation environments. When a Wi-Fi signal spreads through an environment, it undergoes modifications influenced by the structures it encounters, such as walls, persons, and objects. These interactions alter the signal features, including amplitude, phase, and propagation path, shaping it according to the specific properties of the crossed elements. Among the advanced techniques designed to manage these effects, Orthogonal Frequency Division Multiplexing (OFDM)~\cite{5635467} has become a key modulation scheme in wireless communications. OFDM divides the available bandwidth into multiple subcarriers, each carrying a portion of the transmitted signal. This division enables efficient utilization of the frequency spectrum and provides significant mitigation against multipath fading, a phenomenon caused by the interactions between the signal and the environment. A key outcome of the OFDM structure is the ability to analyze the behavior of the channel at the level of individual subcarriers. This is achieved through the Channel Frequency Response (CFR)~\cite{8259000}, which represents the frequency-dependent characteristics of the channel. The CFR captures how the channel modifies the transmitted signal at each subcarrier, encapsulating changes in signal strength, attenuation, and other frequency-selective effects introduced during propagation. To further quantify these modifications and generalize the analysis across multiple packets, the concept of CSI is introduced. The latter extends the CFR representation by encompassing multiple subcarriers and multiple time instances, providing a comprehensive picture of the behavior of the channel. Formally, in the frequency domain, the relationship between the transmitted and received signals can be expressed as:
\begin{equation}
y = H \cdot x + n,
\end{equation}
where $y \in \mathbb{C}^k$ is the vector of the received signal, $H \in \mathbb{C}^k$ is the CFR vector that describes the effect of the channel on each subcarrier, $x \in \mathbb{C}^k$ is the vector of the transmitted signal, and $n \in \mathbb{C}^k$ is the Additive White Gaussian Noise (AWGN)~\cite{10.1109/TPDS.2012.214} at the receiver. The CFR is sampled at the subcarrier level, thus producing discrete measurements of the channel behavior. For an OFDM system with $K$ subcarriers, the CFR for the $k^{\text{th}}$ subcarrier is defined as:
\begin{figure*}[ht]
\centering
\includegraphics[width=0.95\textwidth]{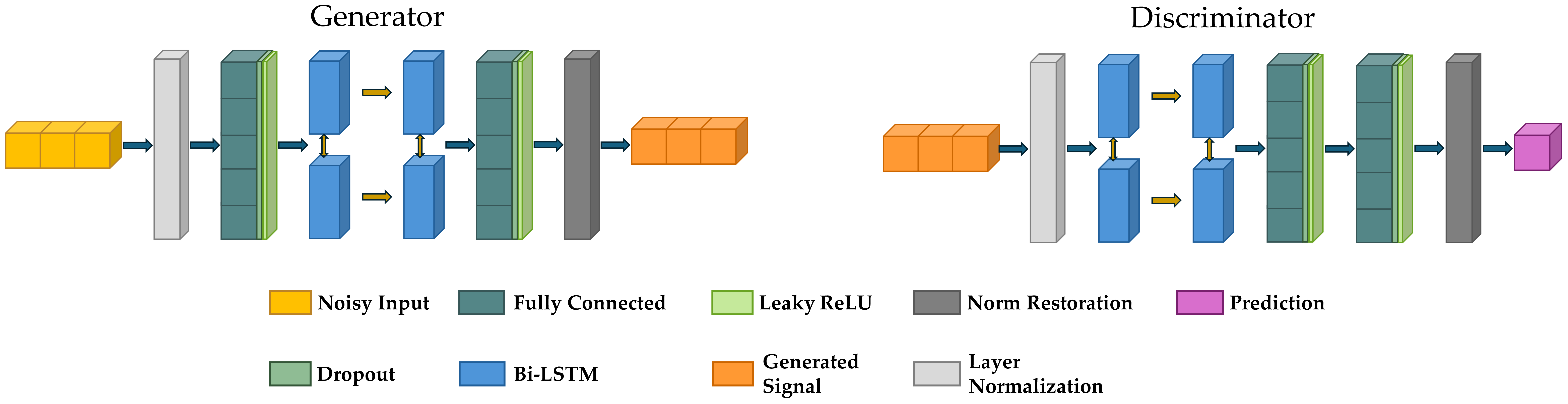}
\caption{Details of the proposed architecture. The first part (left) shows the generator, whose main components are an initial fully connected layer, a Bi-LSTM layer, and a final fully connected layer. The second part (right) shows the discriminator, whose main components are a Bi-LSTM layer and two sequential fully connected layers.}
\label{Figure_03}
\end{figure*}
\begin{equation}
H_k = \frac{Y_k}{X_k},
\end{equation}
where $H_k \in \mathbb{C}$, $Y_k \in \mathbb{C}$, and $X_k \in \mathbb{C}$ represent, respectively, the CFR value, the received signal, and the transmitted signal at the $k^{\text{th}}$ subcarrier. Each CFR value $H_k$ is a complex number expressed as:
\begin{equation}
H_k = |H_k| e^{j\phi_k},
\end{equation}
where $|H_k| \in \mathbb{R^+}$ and $\phi_k \in [0,2\pi)$ represent the amplitude and phase, respectively. For a single transmission packet, the CFR can be represented as a vector:
\begin{equation}
H = \begin{bmatrix} H_1 \\ H_2 \\ \vdots \\ H_K
\end{bmatrix} \in \mathbb{C}^K,
\end{equation}
where $K$ is the total number of subcarriers. When multiple packets are transmitted, the CFR values for all subcarriers across packets form the CSI matrix. For $n$ transmitted packets and $K$ subcarriers the CSI matrix is given by:
\begin{equation}
\mathbf{CSI} =
\begin{bmatrix}
H_{1,1} & H_{1,2} & \dots & H_{1,K} \\
H_{2,1} & H_{2,2} & \dots & H_{2,K} \\
\vdots  & \vdots  & \ddots & \vdots \\
H_{n,1} & H_{n,2} & \dots & H_{n,K}
\end{bmatrix} \in \mathbb{C}^{n \times K},
\end{equation}
where $H_{p,k} \in \mathbb{C}$ is the CFR value for the $k^{\text{th}}$ subcarrier in the $p^{\text{th}}$ packet, and $n$ and $K$ are the total number of transmitted packets and subcarriers, respectively. Observe that the matrix has dimensions $n \times K$ where each row represents the CFR values for a single packet and each column corresponds to a specific subcarrier. Due to the specific focus of this work, a SISO (Single Input Single Output) configuration is adopted, which involves a single transmitting antenna and a single receiving antenna. This choice simplifies the CSI representation, as it avoids the additional complexity introduced by MIMO (Multiple Input Multiple Output) systems, where multiple transmitting and receiving antennas are used. 

By analyzing the base case of SISO, the method concentrates on the core signal cleaning mechanism, thus allowing for a precise evaluation of the proposed digital shielding method. To achieve this, the amplitude component of the CSI is utilized as the primary feature for signal analysis and transformation. Representing the signal strength for each subcarrier, the amplitude encapsulates critical channel properties, including attenuation, multipath reflections, and scattering, which are directly influenced by the environmental interactions of the Wi-Fi signal during propagation. This makes the amplitude a robust abstraction for the channel state, providing a comprehensive and reliable way to represent how the channel behaves. Unlike the phase, which requires precise alignment, the amplitude offers a stable and computationally efficient feature for modeling the channel, preserving the essential characteristics needed for effective signal restoration. Furthermore, by focusing on the amplitude, the proposed method ensures that the restored signal closely approximates its interference-free state, effectively addressing broader signal properties such as phase and spectral coherence. Formally, the amplitude $|H_k| \in \mathbb{R^+}$ for the $k^{\text{th}}$ subcarrier can be defined as follows:
\begin{equation}
|H_k| = \sqrt{\text{Re}(H_k)^2 + \text{Im}(H_k)^2},
\end{equation}
where $\text{Re}(H_k)$ and $\text{Im}(H_k)$ denote the real and imaginary parts of $|H_k|$, respectively. The extracted amplitudes, corresponding to both shielded and unshielded conditions, serve as the basis for training the RaGAN-based architecture.

\subsubsection{Generator and Discriminator Networks:}
\label{generator_and_discriminator_networks}
the initial idea behind the proposed digital shielding method took inspiration from GANs, which are applied to signal denoising in a wide range of application domains, such as ElectroEncephaloGraphy (EEG)~\cite{10131991}, Magnetic Resonance Imaging (MRI)\cite{10350957}, Sound Recognition (SR)\cite{10603113}, and many others. These operational scenarios highlight the versatility of GANs in addressing noise-related challenges across different areas. However, traditional GANs are known to suffer from diverse issues\cite{10.1145/3446374,10.1145/3459992,10.1145/3439723}, including instability during training, generally caused by the balance required between the generator and discriminator, slow convergence due to the iterative nature of adversarial optimization, and challenges in processing signals with complex dynamic patterns that demand a deep understanding of temporal dependencies. To overcome these limitations, this work proposes the adoption of a RaGAN, a variant specifically designed to address the shortcomings of conventional GANs. In general, the RaGAN modifies the objective of the discriminator to ensure a more stable training process, thus improving both convergence speed and performance. In particular, RaGAN introduces a relativistic average discriminator, which considers the real and generated samples as relative to each other rather than absolute entities. This approach aims to provide a more balanced and robust training process by incorporating the notion of relativistic comparison.

In Fig.~\ref{Figure_03}, details of the implemented generator and discriminator in the proposed RaGAN-based architecture are presented. As well known, traditional GANs designed for signal denoising consist of two primary components, i.e., a generator and a discriminator. These networks collaborate within an adversarial process to improve the quality of noisy input signals. Specifically, the generator is trained to consider every environment information contained in the CSI amplitude data as noise and transform signals acquired from real-world environments into their denoised counterparts by learning the mapping between noisy and clean signal domains. Through adversarial training, the generator iteratively enhances its capability to produce outputs that closely approximate reference clean signals, which represent the ground truth of the environment-free spectrum. The discriminator is tasked with distinguishing between real clean signals and those generated by the generator. By evaluating the quality of the generated outputs and providing constructive feedback, the discriminator plays a crucial role in enabling the generator to progressively refine its ability to create realistic and accurate denoised signals. Once the training process is complete, the generator operates independently to process new noisy signals, reconstructing their clean spectra with high fidelity. This separation allows the trained generator to function as a standalone model for effective signal denoising, capable of extracting and restoring spectral information from noisy inputs.

From an overall perspective, the proposed RaGAN-based architecture follows the same logical pipeline as the previously described GAN-based architecture for signal denoising but differs from both conventional GANs and standard RaGANs in several fundamental aspects. To begin with, differently from conventional GANs, the introduction of a relativistic average discriminator allows the proposed RaGAN-based architecture to overcome the limitations outlined earlier. By evaluating the relative likelihood that real samples appear more authentic compared to generated ones, instead of relying on absolute classifications, the relativistic approach reduces instability during training. This comparative mechanism minimizes sensitivity to imbalances between the generator and discriminator, thus fostering a more stable optimization process. Additionally, the relativistic discriminator enhances convergence by providing richer informative gradients during adversarial training. Lastly, this comparative approach enables the model to better handle complex dynamic patterns by concentrating on the relative distinctions between noisy and clean signals, thus enhancing the ability of the generator to adapt to complex temporal and spectral variations in the input data. In addition, unlike standard RaGANs, the proposed architecture integrates Bi-LSTM networks as the core computational units for both the generator and discriminator. This modification introduces significant advantages as Bi-LSTMs are specifically designed to capture both forward and backward temporal dependencies in sequential data, thus making them well-suited for analyzing the waveform and time-dimension characteristics of noisy signals. By processing the signal bidirectionally, Bi-LSTMs enable the architecture to extract contextual information from both past and future signal states, providing a comprehensive understanding of the input.

Following the architecture presented in Fig.~\ref{Figure_03} (left), the generator is designed to transform noisy amplitudes extracted from CSI into clean, interference-free counterparts. The input to the generator is represented by a matrix $X \in \mathbb{R}^{n \times K}$, where $n$ is the number of packets and $K$ is the number of subcarriers. The initial stage of the generator consists of a fully connected layer that maps the input from its original dimensionality, $\mathbb{R}^{n \times K}$, into a latent space, $\mathbb{R}^{n \times d}$, where $d$ is a predefined dimension. This projection is defined as follows:
\begin{equation}
X' = \psi(W_1 \cdot X + b_1),
\end{equation}
where $W_1$ and $b_1$ represent the weight matrix and bias vector, respectively, and $\psi(.)$ is a nonlinear activation function designed to introduce a small slope in the negative region to prevent gradient saturation (i.e., Leaky ReLU). The latent representation $X'$ is then processed by a Bi-LSTM layer, which captures both forward and backward temporal dependencies. For each timestep $t$, the forward and backward hidden states, $\overrightarrow{h_t}$ and $\overleftarrow{h_t}$, are concatenated to form the complete hidden state:
\begin{equation}
h_t = [\overrightarrow{h_t}, \overleftarrow{h_t}] \in \mathbb{R}^{2d}.
\end{equation}
Subsequently, the Bi-LSTM processes the sequence to generate an output matrix:
\begin{equation}
H = [h_1, h_2, \dots, h_n] \in \mathbb{R}^{n \times 2d},
\end{equation}
which is passed through a final fully connected layer to map the data back to its original dimensions:
\begin{equation}
Y' = \sigma(W_2 \cdot H + b_2),
\end{equation}
where $W_2 \in \mathbb{R}^{K \times 2d}$ and $b_2 \in \mathbb{R}^K$ are the weight matrix and bias vector of the concluding fully connected layer, and $\sigma(.)$ is a nonlinear activation function designed to map input values to the range $[0,1]$, making it suitable for representing probabilities and normalizing outputs (i.e., Sigmoid).

Following the architecture presented in Fig.~\ref{Figure_03} (right), the discriminator is designed to evaluate the authenticity of the amplitudes generated by the generator. Its role is to differentiate between clean amplitudes from the ground truth and those synthesized by the generator. This is achieved through a Bi-LSTM network complemented by two sequential fully connected layers. The discriminator takes as input the same matrix $X \in \mathbb{R}^{n \times K}$, representing either real or generated amplitudes. The Bi-LSTM processes the temporal structure of $X$ and outputs a matrix $H \in \mathbb{R}^{n \times 2d}$, analogous to the hidden states of the generator. The first fully connected layer reduces the dimensionality:
\begin{equation}
H' = \psi(W_3 \cdot H + b_3),
\end{equation}
where $W_3 \in \mathbb{R}^{d \times 2d}$ and $b_3 \in \mathbb{R}^d$. The second fully connected layer computes the final output:
\begin{equation}
z = \sigma(W_4 \cdot H' + b_4),
\end{equation}
where $W_4 \in \mathbb{R}^{1 \times d}$ and $b_4 \in \mathbb{R}^1$. The output $z$ is a scalar indicating the probability that the input belongs to the real data distribution. 

It is important to emphasize that the functionality of the Bi-LSTM differs between the generator and the discriminator. In the generator, Bi-LSTM is employed to capture temporal correlations within noisy inputs, thus enabling the network to reconstruct clean amplitudes that closely approximate their interference-free counterparts. This process involves learning a mapping that adapts to variations introduced by environmental noise, thereby ensuring effective denoising. Conversely, in the discriminator, Bi-LSTM is utilized to differentiate temporal patterns between real (i.e., clean) and generated (i.e., denoised) signals. This is achieved by analyzing the sequential structure of the input, where bidirectional processing enhances the ability of the network to detect inconsistencies in the generated samples.
\begin{figure*}[ht]
\centering
\includegraphics[width=0.78\textwidth]{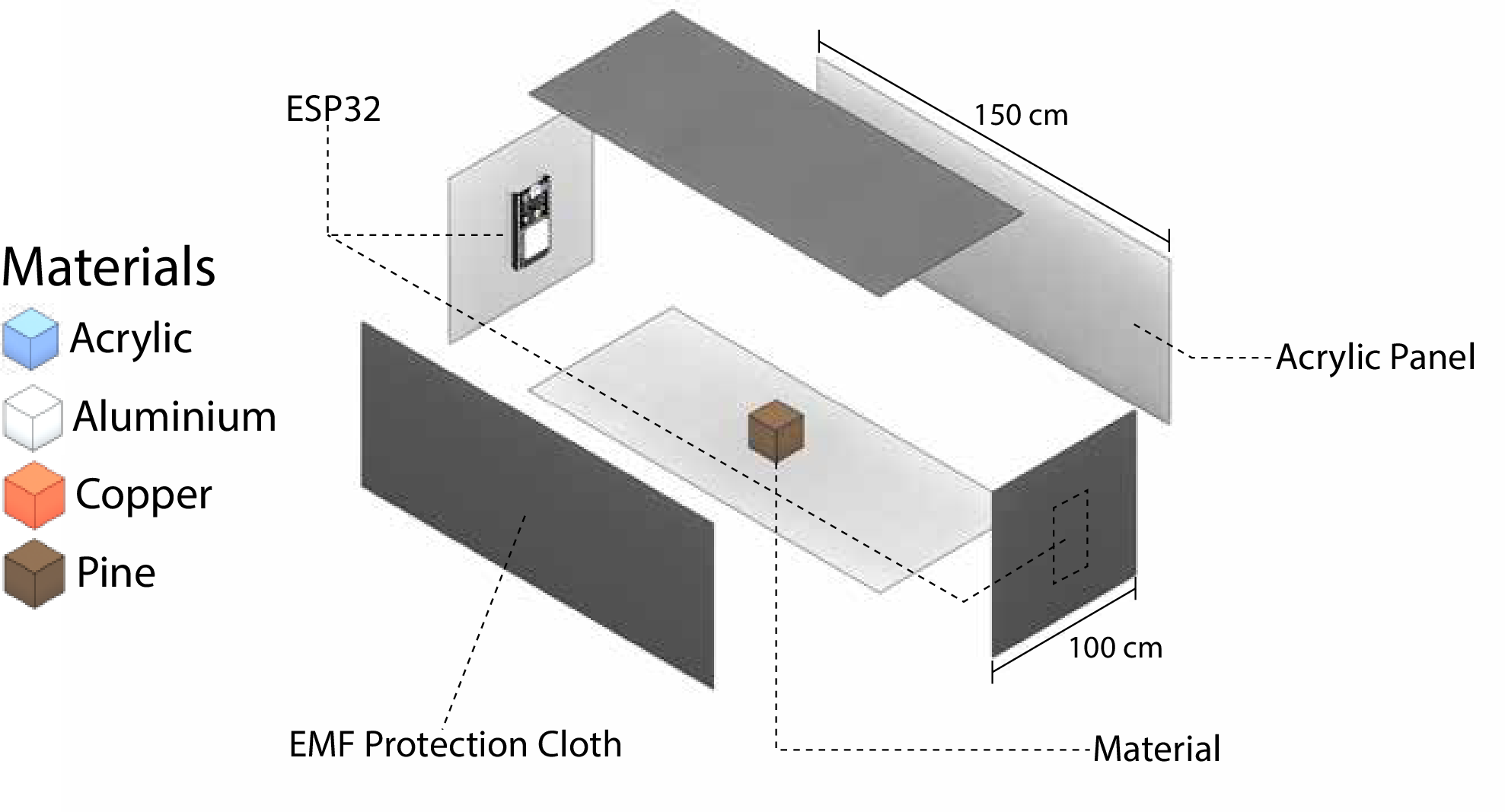}
\caption{Acrylic box lined with electromagnetic shielding fabric, designed to replicate the effects of a Faraday cage. The box isolates objects, enabling the RaGAN to learn the impact of physical shielding. The legend shows four cubes made of different materials (acrylic, aluminum, copper, and pine), each with dimensions of $2 \times 2 \times 2$ cm, which were used in the classification task to validate the effectiveness of the proposed denoising method. Two ESP32 devices are positioned within the box, serving as the transmitter and receiver for Wi-Fi signals.}
\label{Figure_04}
\end{figure*}

A final innovation introduced in the proposed model concerns the loss functions. Specifically, while the discriminator employs the relativistic average loss, a standard choice for RaGAN-based architectures, the loss function of the generator has been customized. Inspired by the work of Ledig et al.\cite{8099502}, the proposed model incorporates an overall loss function for the generator that combines content loss and contrastive loss. Considering both aspects, the generator is guided by both content information and confrontation between signal patterns. Formally, the customized loss function is expressed as follows:
\begin{equation}
L_G = L_c + \lambda L_G^{Ra},
\end{equation}
where $L_c$ and $L_G^{Ra}$ represent the content loss and confrontation loss, respectively, and $\lambda$ is the coefficient balancing the contributions of these loss components. By jointly optimizing the content and confrontation losses, the generator effectively restores noisy signals $s$ to their noise-free counterparts $s_r$. The content loss function, $L_c$, is made up of two components, i.e., $L_{MSE}$ and $L_1$, more specifically: 
\begin{equation}
L_c = \frac{(L_{MSE} + L_1)}{2},
\end{equation}
where $L_{MSE}$ and $L_1$ quantify the Mean Square Error (MSE) and the Absolute Error (AE) between the denoised amplitude $G(s)$ and the clean amplitude $s_r$, respectively. Note that $G(s)$ is the output of the generator and $n$ denotes the length of a single sample. Formally, the two measures are expressed as follows: 
\begin{equation}
L_{MSE} = \frac{1}{n} \sum_{i=1}^{n} (s_{r_i} - G(s)_i)^2,
\end{equation}
\begin{equation}
L1 = \frac{1}{n} \sum_{i=1}^{n} |s_{r_i} - G(s)_i|.
\end{equation}

\subsection{Multi-Class SVM}
\label{multi-class_SVM}
Through iterative adversarial training, the RaGAN enhances the ability of the generator to produce amplitudes that closely resemble their ground truth environment-free representations. Following training, the generator is deployed as a digital shielding mechanism, actively denoising real-world signal amplitudes on-the-fly, restoring them to their domain-free forms. To evaluate the effectiveness of the proposed method, a practical and challenging task was selected, i.e., material classification. This task involves distinguishing between four representative materials, i.e., acrylic, aluminum, copper, and pine, by analyzing the unique modifications induced in Wi-Fi signals as they propagate through the materials of varying compositions. Note that the four materials can be considered highly representative of the objects and items commonly present in real-world environments.

To evaluate the performance of the proposed domain adaptation method, a multi-class SVM classifier was employed. This choice was motivated by two key considerations. On one hand, the SVM is generally considered as a robust and reliable classifier. On the other hand, it is known that its performance can degrade when dealing with particularly complex, non-linear, and noisy data. Given the nature of the Wi-Fi signal amplitudes, characterized by significant interference and a high degree of disorder, a poorly performing denoising method would have resulted in low classification accuracy. However, as demonstrated in the experimental section, the classification accuracy achieved in the material classification task remains remarkably high, despite the complexity of the input.

\section{Experimental Setup and Results}
\label{experimental_setup_and_results}
This section shows the main stages involved in obtaining the experimental results. The first describes the construction of the shielded box and the process of acquiring the data used for training the RaGAN-based architecture and testing the multi-class SVM. The second details the training of the architecture and presents the accuracy achieved in the material classification case study, demonstrating the effectiveness of the domain adaptation method.

\subsection{Shielded Box and Data Collection}
\label{shielded_box_and_data_collection}
As shown in Fig.~\ref{Figure_04}, a shielded box was constructed to replicate the effects of a Faraday cage. The structure, made of acrylic due to its mechanical properties, has internal dimensions measuring $150 \times 100 \times 100$ cm. The external surfaces of each panel of the box were lined with high-performance electromagnetic shielding fabric, designed to minimize external radio frequency interference and ensure a fully isolated, noise-free internal environment for signal acquisition. To facilitate measurements, the box was designed with the removable upper panel, thus allowing for the easy placement and adjustment of objects inside. Additionally, two openings were placed on opposite sides of the box to accommodate two ESP32 Wi-Fi-enabled microcontrollers\cite{9900419}. One device was used for signal transmission and the other for signal reception.

Regarding ESP32, it is a highly versatile device known for its dual-core processors and integrated Wi-Fi capabilities. Operating at 2.4 GHz and supporting the IEEE 802.11n standard, the ESP32 is configured in this work with a SISO setup involving one transmitting and one receiving antenna. This configuration simplifies the representation of CSI while preserving the essential characteristics of the signal. Leveraging the OFDM modulation scheme, a key feature of the IEEE 802.11n standard, the ESP32 divides the available bandwidth into multiple subcarriers, thus enhancing robustness against multipath fading and interference and ensuring data integrity during transmission. In the SISO configuration, the ESP32 achieves a theoretical maximum transmission speed of 150 Mbps with a 40 MHz channel width. This capability makes it well-suited for Wi-Fi signal denoising tasks, providing high spectral efficiency and enabling precise extraction of channel properties, including amplitude. In the experimental setup, one ESP32 device functions as a transmitter, sending Wi-Fi packets at a constant rate of 100 packets per second to create a stable signal source. The transmitted signal interacts with the object placed inside the shielded box, and the resulting signal is received by the second ESP32 device configured as a receiver. The receiver records the CSI, capturing variations in amplitude introduced by the material properties of the object during signal propagation. To ensure high-quality data collection, the ESP32 receiver is configured to extract CSI data from 64 subcarriers, thus providing a detailed frequency-domain analysis of the channel. Both ESP32 devices are synchronized during acquisition to maintain consistency in packet transmission and reception. This setup ensures accurate and reproducible measurements of Wi-Fi signals under both shielded and unshielded conditions. The modular design of the box and the adaptability of the ESP32 devices contribute to a controlled and flexible experimental environment, enabling the reliable acquisition of clean Wi-Fi signal data for various test scenarios. In Table~\ref{tab_EPS32_par}, the main parameters selected for configuring the device settings are reported.
\begin{table}[ht]
\caption{ESP32 - device settings} \label{tab_EPS32_par}
{\begin{tabular}{@{}lcccr@{}}
\hline
Parameter                 & Transmitter (Tx)/Receiver (Rx)    \\ \hline
Wi-Fi Standard            & IEEE 802.11n                      \\
Modulation                & OFDM                              \\
Frequency                 & 2.4 GHz                           \\
Configuration             & SISO                              \\
Bandwidth                 & 40 MHz                            \\
N° of Packets             & 100/s                             \\
N° of Subcarriers         & 64                                \\  
Acquisition Time          & 10 s                              \\   
Output Data               & N/A(Tx)/Raw-CSI(Rx)               \\
Synchronization           & Clock(Tx)/With-Tx(Rx)             \\
\hline
\end{tabular}}
\end{table}

As previously mentioned, each Wi-Fi packet transmitted under the IEEE 802.11n standard consists of 64 subcarriers. Of these, 52 are actively used for data transmission, while the remaining subcarriers, categorized as pilot and guard subcarriers, serve as reference signals. Pilot subcarriers assist with channel estimation and phase correction, ensuring synchronization and stability during transmission. Guard subcarriers, positioned at the edges of the frequency spectrum, help mitigate inter-channel interference and prevent spectral leakage. Together, these unused subcarriers contribute to the overall robustness and reliability of the communication channel.

Now that the acquisition specifications, such as the number of packets and subcarriers, have been detailed, it is possible to quantify the network configuration to facilitate the reproducibility of the proposed method. Referring to Fig.~\ref{Figure_03} (left), the generator consists of two fully connected layers and a Bi-LSTM layer, designed to effectively capture both the waveform and time-dimension characteristics of the noisy signal. The input and output dimensions of the generator are both set to $1000 \times 52$. Initially, the input tensor undergoes normalization, after which it is passed through a fully connected layer with an input size of $52$ and an output size of $256$. To enhance generalization, a dropout layer with a probability of $0.3$ is applied to the tensor. Subsequently, a Leaky ReLU activation function is introduced to add non-linearity. The tensor is then processed by a Bi-LSTM layer with an input size of $256$ and a hidden size of $256$. Since the Bi-LSTM processes the input bidirectionally, it produces an output with a dimension of $512$. The output of the Bi-LSTM is then fed into another fully connected layer with an input size of $512$ and an output size of $1000 \times 52$. Another dropout layer, with a dropout probability of $0.3$, is applied to this tensor, followed by another Leaky ReLU activation. Finally, the tensor undergoes layer normalization to ensure numerical stability and better convergence. Referring to Fig.~\ref{Figure_03} (right), the discriminator network consists of two fully connected layers and a Bi-LSTM layer. Initially, the input tensor undergoes layer normalization. It is then passed through a Bi-LSTM layer with an input size of $52$ and a hidden size of $256$. Following the Bi-LSTM, the output corresponding to the last time step is processed by a fully connected layer with an input size of $512$ (i.e., bidirectional output size) and an output size of $256$. A dropout layer, with a probability of $0.3$, is applied to the tensor, followed by a Leaky ReLU activation to introduce non-linearity. The tensor is then passed through another fully connected layer with an input size of $256$ and an output size of $1$, representing the probability of the input being real or fake. 

The dropout rate of $0.3$ was selected based on established best practices in literature\cite{10.5555/2627435.2670313,10.5555/3042817.3043055}, particularly for architectures designed to handle high-dimensional data like Wi-Fi signal amplitudes. This value provided a good balance between preventing overfitting and retaining critical information for effective learning. Initial experiments with alternative dropout rates, e.g., $0.2$, $0.4$, demonstrated that $0.3$ offered the most consistent performance in terms of convergence and generalization on validation data.

An ad-hoc dataset was collected to train the RaGAN-based architecture and evaluate its signal denoising capabilities using a multi-class SVM. Four objects with identical sizes of $2 \times 2 \times 2$ cm but composed of different materials, i.e., acrylic, aluminum, copper, and pine, were acquired following the parameters specified in Table~\ref{tab_EPS32_par}. The objects were positioned at the center of the box, as shown in Fig.~\ref{Figure_04}, and acquisitions were performed twice. The first set of acquisitions was conducted inside the shielded box, thus ensuring noise-free conditions, while the second set was performed with all the shielding panels removed, maintaining the same distance between devices, thus reflecting real-world conditions with noise. As a result, two datasets were obtained, $D_S$, i.e., shielded dataset, containing the spectra of the four objects under shielded conditions, and $D_U$, i.e., unshielded dataset, containing the spectra of the same objects under unshielded conditions. More specifically, inside the shielded box, each of the four objects was acquired $30$ times, along with $30$ background noise recordings without any objects present. Similarly, outside the shielded box, $30$ acquisitions were performed for each object and $30$ for the background noise. To ensure generalization, particularly in the noisy environment outside the box, acquisitions were conducted on different days and at varying times. In total, $300$ acquisitions were collected, with $150$ obtained under shielded conditions and $150$ in real-world noisy conditions. It is important to note that the use of only four materials for training and validation should not be seen as a limitation. First, the selected materials can be considered highly representative of the objects and items commonly found in real-world environments. Second, the task of recognizing the material composition of an object using Wi-Fi signals that pass through it is inherently challenging and complex, and remains an area still underexplored in the current literature. 

\subsection{RaGAN Training and Classification}
\label{ragan_training_and_classification}
Once the datasets $D_S$ and $D_U$ were prepared, each consisting of $150$ files (i.e., $30$ per object and $30$ for background noise), the RaGAN model was trained on both to set and refine the ability of the generator to denoise Wi-Fi signals. To further evaluate the model performance, experiments were conducted using two distinct versions of the datasets. The first version included the complete datasets, consisting of all $150$ files from each set. The second set of experiments utilized subsets of the datasets, each comprising data for all objects and background noise, but limited to acquisitions performed on a single day. Over a span of ten days, each day included the acquisition of all four objects and background noise both inside and outside the shielded box, resulting in a total of $30$ acquisitions per day. This setup allowed for a detailed examination of the model ability to generalize across temporal variations while maintaining a robust evaluation framework. Based on the described setup, a grid search approach~\cite{10.5555/2188385.2188395} was employed to optimize the hyperparameters of the RaGAN model and assess its performance across different temporal subsets. The ranges of the hyperparameters explored during this process and the optimal values selected after evaluation on the validation set are summarized in Table~\ref{tab:gridsearch} and Table~\ref{tab:optimal}, respectively. In both tables, $G$ and $D$ are used to denote the generator and discriminator components of the proposed model.
\begin{table}
\caption{Grid search hyperparameter ranges} \label{tab:gridsearch}
{\begin{tabular}{@{}ll@{}}
\hline
\textbf{Hyperparameter} & \textbf{Range} \\
\hline
Batch size          & \{5, 10, 15, 30\} \\
Dropout & \{0.2, 0.3, 0.4\}  \\
Learning rate (G)   & \{0.0001, 0.001, 0.01, 0.1\} \\
Learning rate (D)   & \{0.0001, 0.001, 0.01, 0.1\} \\
Optimizer           & \{AdamW\} \\
$\lambda$           & \{10, 50, 100, 500\} \\
\hline
\end{tabular}}
\end{table}
\begin{table}
\caption{Optimal hyperparameter set} \label{tab:optimal}
{\begin{tabular}{@{}lp{0.45\columnwidth}@{}}
\hline
\textbf{Hyperparameter} & \textbf{Value} \\
\hline
Batch size          & 5 \\
Dropout & 0.3 \\
Learning rate (G)   & 0.001 \\
Learning rate (D)   & 0.0001 \\
Optimizer           & AdamW \\
$\lambda$           & 100 \\
\hline
\end{tabular}}
\end{table}
\begin{figure}
\centering
\includegraphics[width=3.2in]{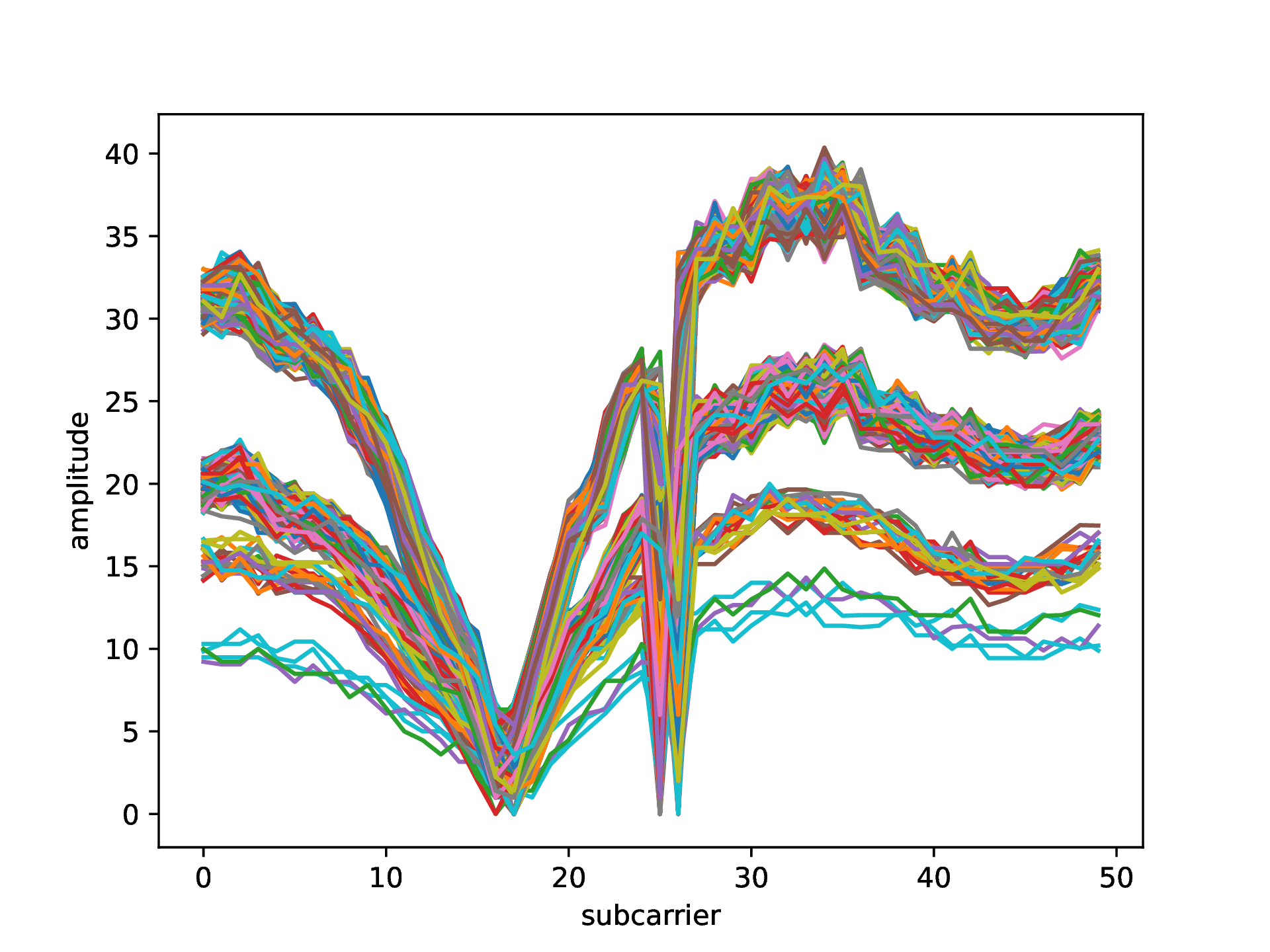}
\caption{Amplitudes extracted from the CSI: the plot shows the reconstructed spectrum of the copper cube derived from data acquired in a real-world environment.}
\label{Figure_05}
\end{figure}

The main objective of applying the RaGAN model was to create a new dataset $D'_S$ of reconstructed and denoised signals, derived from the unshielded dataset $D_U$. The effectiveness of RaGAN in achieving this goal is demonstrated through both qualitative and quantitative analyses using a multi-class SVM classifier. In Fig.~\ref{Figure_05}, the example shown in Fig.~\ref{Figure_02} is continued, where the amplitudes extracted from the CSI for the copper cube were displayed in the unshielded real-world environment (top) and the shielded environment (bottom), with the latter representing the ground truth spectrum obtained in a noise-free setting. Specifically, Fig.~\ref{Figure_05} presents the reconstructed amplitude for the same copper cube acquired in the unshielded environment illustrated in Fig.~\ref{Figure_02} (top). The results clearly demonstrate that the spectra reconstructed by RaGAN closely match the ground truth, accurately reflecting the key features and attenuation patterns linked to the shielded environment. This emphasizes the capability of the model to produce a denoised dataset, $D'_S$, that effectively represents the shielded spectrum.
\begin{figure}
    \centering
    \includegraphics[width=3.2in]{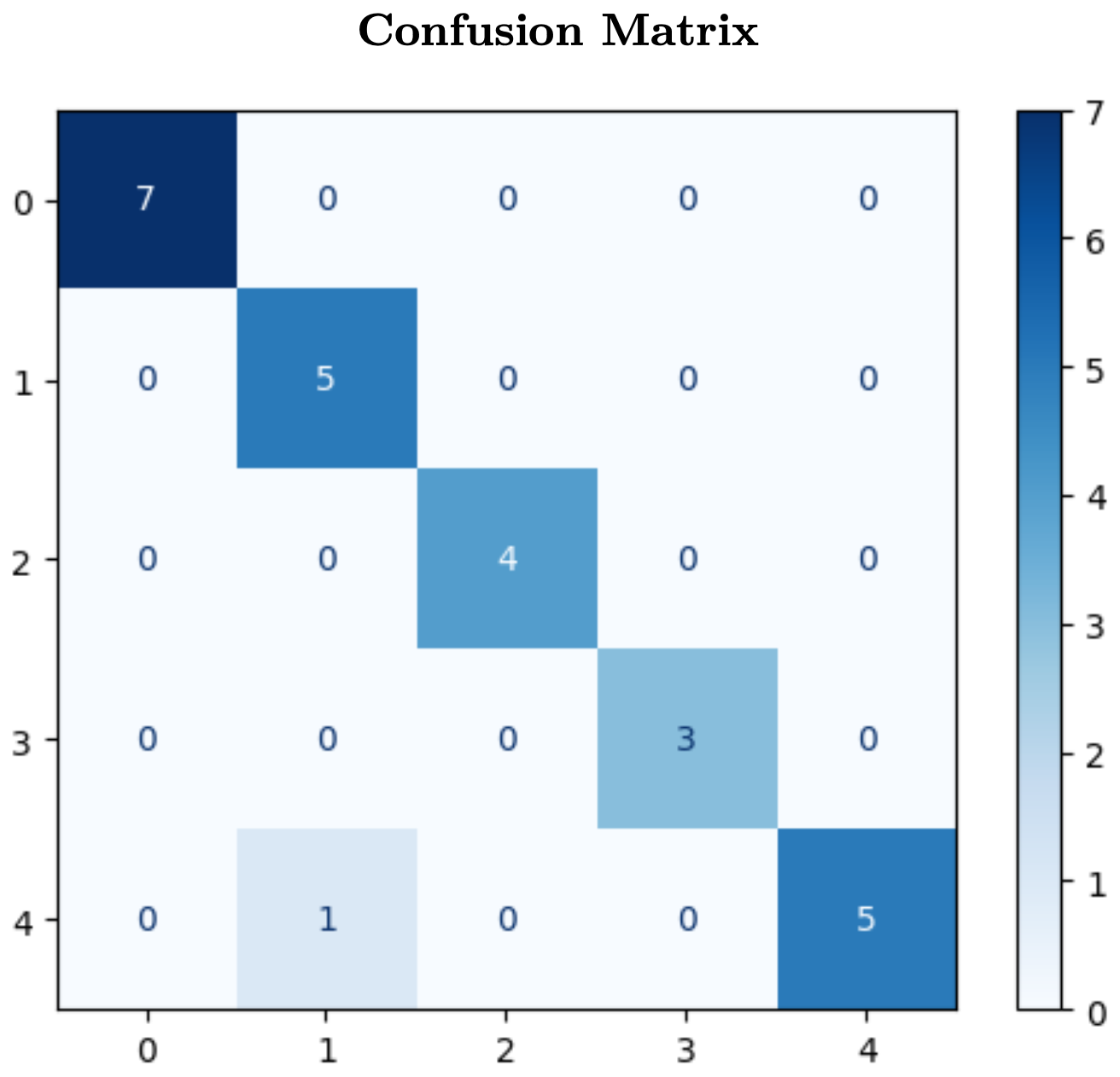}
    \caption{SVM classifier: confusion matrix.}
    \label{confusion_matrix}
\end{figure}

To quantitatively evaluate the synthesized spectra generated by the RaGAN model, a multi-class SVM classifier was employed. The classification task aimed to distinguish between spectra corresponding to different materials, each one characterized by unique spectral features influenced by its physical and electromagnetic properties. The classifier was trained using both the shielded spectra, serving as the ground truth, and the unshielded spectra as input data. This evaluation approach simultaneously assesses the similarity between the generated spectra and the shielded spectra while also evaluating the capacity of the model to retain the distinct features that differentiate materials. The multi-class SVM was initially trained on the shielded spectra to learn material-specific features, e.g., attenuation patterns and frequency responses, that are unique to each material. Note that these features allow the classifier to differentiate between spectra corresponding to different materials. The trained classifier was subsequently tested on the spectra reconstructed by the RaGAN generator, achieving an impressive accuracy score of 96\%. This result demonstrates that the proposed method effectively removes noise from the unshielded input while preserving the distinguishing characteristics that make each material identifiable. The confusion matrix, presented in Fig. 6, further illustrates the classification performance, highlighting high accuracy in differentiating the reconstructed spectra. Most classifications align correctly along the main diagonal, thus confirming the ability of the RaGAN-generated spectra to effectively retain the distinctive features of the specific materials. In addition to the classification-based evaluation, a Mean Squared Error (MSE)\cite{rainioEvaluationMetricsStatistical2024} analysis was conducted to quantify the pointwise similarity between the signals reconstructed by the RaGAN model and the corresponding ground truth spectra acquired in the shielded environment. The reported MSE value of approximately 0.19 (normalized over a 0-to-1 scale), computed as an average across all ground truth and reconstructed signal pairs obtained during experimentation, was calculated after normalizing the signal amplitudes on a scale from zero to one. This normalization ensures a consistent evaluation metric across all samples and facilitates fair comparisons between different signals. While MSE offers a useful numerical indication of the overall reconstruction quality, it does not necessarily capture the preservation of discriminative spectral features that are essential for classification tasks. In Wi-Fi sensing scenarios, such as the material identification task proposed in this study, the primary objective is not only to reduce noise but also to preserve semantic information that enables class separability. In this regard, the classification accuracy achieved by the SVM, 96\% on the reconstructed spectra, demonstrates that the denoised signals retain the critical features required to distinguish among different materials. Taken together, the MSE evaluation, classification accuracy, and visual inspection of the signal profiles provide a multi-faceted and robust assessment of the performance of the model. This integrated evaluation framework confirms the ability of the proposed method to effectively suppress environment-specific interference while preserving the information necessary for downstream classification.

\subsection{Ablation Study and Analysis} 
\label{ablation_study_and_analysis} 
The final architecture of our model was carefully selected through a thorough process of empirical testing, which involved exploring a range of different combinations of layers and generative techniques. We began by evaluating various configurations of Bi-LSTM layers, exploring different possibilities, including setups with and without linear layers for data processing. This systematic approach allowed us to gather valuable performance data, which is summarized in Table~\ref{tab:ablation}. Throughout our testing, we found that the generative Bi-LSTM models exhibited remarkably similar performance, whether they employed only LSTM layers or utilized a combination of a single LSTM layer along with linear layers. Given the significant increase in the number of parameters required when using only LSTM layers, we made the informed decision to adopt the lighter model. This model not only proved to be less complex but also demonstrated performance metrics that were comparable to those of the more parameter-heavy configurations. In addition to our analysis of the Bi-LSTM models, we also explored variations alongside the RaGAN method, examining the performance of the CGAN approach through a similar analytical framework. The results revealed that the Bi-LSTM\_CGAN achieved levels of performance that were comparable to those of the Bi-LSTM\_RaGAN. This finding demonstrated that linear layers can perform effectively, similarly to LSTM layers, while potentially simplifying the architecture. To provide a comprehensive understanding of our experimental framework, we also presented the baseline models utilized in our analysis. These included a Denoising Autoencoder (DAE) and a SVM, both of which were applied to the preprocessed noisy data collected during the testing phase. The DAE represents a straightforward implementation, consisting of four hidden layers in both the encoder and decoder components, allowing for effective denoising of the input data. On the other hand, the SVM used in our experiments was consistent with the one leveraged for classifying the generated signals produced by the GAN models. This SVM was based on a Gaussian radial basis function, with a specified parameter of $\gamma = 1/2\sigma^2$, enabling robust classification performance within the context of our analyses. By assessing these various models, we were able to gain deeper insights into their effectiveness and suitability for the tasks at hand.
\begin{table}
\caption{Ablation Study} 
\label{tab:ablation}
{\begin{tabular}{@{}lp{0.45\columnwidth}@{}}
\hline
\textbf{Model} & \textbf{Accuracy} \\
\hline
Bi-LSTM\_RaGAN / w linear   & 0.96 \\
Bi-LSTM\_RaGAN / wo linear  & 0.96 \\
Bi-LSTM\_CGAN / w linear   & 0.92 \\
Bi-LSTM\_CGAN / wo linear  & 0.92 \\
SVM                       & 0.72 \\
DAE                       & 0.88 \\
\hline
\end{tabular}}
\end{table}

\section{Conclusion}
\label{conclusion}
To the best of our knowledge, this paper introduces, for the first time in the literature, a RaGAN-based architecture with customized Bi-LSTM-based generator and discriminator networks, specifically designed to denoise Wi-Fi signals by digitally replicating the effects of a Faraday cage. The effectiveness of this foundational study on cross-domain adaptation is demonstrated through a challenging material classification task. Wi-Fi signals of four objects made of distinct materials (i.e., acrylic, aluminum, copper, and pine) were acquired in real-world scenarios and subsequently purged of environment-specific information using the proposed architecture. A multi-class classifier, trained on interference-free counterparts acquired within the shielded box, was employed to classify the denoised signals, achieving an remarkable accuracy of 96\%. This result underscores the ability of the domain adaptation and denoising mechanism to effectively restore signal fidelity. Furthermore, the evaluation highlights the added value of the proposed approach, as the high classification accuracy demonstrates its potential for advanced security applications. By reliably distinguishing materials based on the denoised spectra, the method provides an innovative solution for identifying the nature and composition of objects, including those that may be carried by individuals in sensitive or high-security environments. In future work, a more comprehensive dataset will be collected, including a broader variety of materials. The proposed model will also be evaluated in more complex real-world environments to assess its generalizability. Finally, while this study adopted a multi-class SVM to validate the effectiveness of the signal purification process, future efforts will explore advanced classification frameworks specifically tailored to material recognition tasks. These may include state-of-the-art deep learning models, such as Neural Dynamic Classification (NDC) algorithms\cite{7990591}, Dynamic Ensemble Learning (DEL) techniques\cite{10.1007/s00521-019-04359-7}, Finite Element Machines (FEM) for fast learning\cite{FEMa}, and self-supervised learning approaches\cite{9837871,CWE}.

\section*{Acknowledgement}
This work was supported by ``EYE-FI.AI: going bEYond computEr vision paradigm using wi-FI signals in AI systems'' project of the Italian Ministry of Universities and Research (MUR) within the PRIN 2022 Program (Grant Number: 2022AL45R2) (CUP: B53D23012950001) and MICS (Made in Italy – Circular and Sustainable) Extended Partnership and received funding from Next-Generation EU (Italian PNRR – M4 C2, Invest 1.3 – D.D. 1551.11-10-2022, PE00000004). CUP MICS B53C22004130001.

\bibliography{References}
\end{document}